\pdfoutput=1
\documentclass[preprint, authoryear]{elsarticle}
\journal{Journal of Computers and Geosciences}

\usepackage[scriptsize, TABBOTCAP]{subfigure}
\usepackage{etoolbox}
\usepackage{verbatim}
\usepackage{graphicx}
\usepackage{multirow}
\usepackage{rotating}
\usepackage{xcolor}
\usepackage{lineno}
\usepackage{natbib}
\usepackage{amsmath}
\usepackage{amsfonts}

\RequirePackage[normalem]{ulem} 
\RequirePackage{color}\definecolor{RED}{rgb}{1,0,0}\definecolor{BLUE}{rgb}{0,0,1} 

\usepackage{siunitx}
\usepackage[acronym]{glossaries} 
\setacronymstyle{long-short}
\newacronym{GA}{GA}{Genetic Algorithm}
\newacronym{AnEn}{AnEn}{Analog Ensemble}
\newacronym{ELBO}{ELBO}{Evidence Lower Bound}
\newacronym{VAE}{VAE}{Variational Autoencoder}
\newacronym{CVAE}{CVAE}{Conditional Variational Autoencoder}
\newacronym{KL}{KL}{Kullback-Leibler}
\newacronym{FLT}{FLT}{Forecast Lead Time}
\newacronym{EA}{EA}{Evolutionary Algorithm}
\newacronym{RMSE}{RMSE}{Root Mean Square Error}
\newacronym{NWP}{NWP}{Numerical Weather Prediction}
\newacronym{NAM}{NAM}{North American Mesoscale Forecast System}
\newacronym{DOUG}{DOUG}{Dynamically Optimized Unstructured Grid}
\newacronym{OMEGA}{OMEGA}{Operational Multiscale Environment Model with Grid Adaptivity}
\newacronym{ECMWF}{ECMWF}{European Centre for Medium-Range Weather Forecasts}
\newacronym{PDF}{PDF}{Probability Density Function}

\bibpunct[, ]{(}{)}{;}{a}{}{,}

\makeatletter
\def\blfootnote{\xdef\@thefnmark{}\@footnotetext}
\makeatother

\patchcmd{\pprintMaketitle}
  {\hrule\vskip12pt}
  {\hrule\vskip12pt\ifvoid\extrainfobox\else\unvbox\extrainfobox\par\vskip12pt\fi}
  {}{}

\newsavebox\extrainfobox

\begin{document}
\begin{frontmatter} 
\title{Probabilistic Forecasting using Deep Generative Models}

\author[NCAR]{Alessandro Fanfarillo}\corref{correspondingAuthor}
\cortext[correspondingAuthor]{Corresponding author.}
\ead{elfanfa@ucar.edu}

\author[IOWA]{Behrooz Roozitalab}

\author[PennState]{Weiming Hu}
\ead[url]{http://geoinf.psu.edu/}

\author[PennState,NCAR]{Guido Cervone}

\address[NCAR]{Research Applications Laboratory, National Center for Atmospheric Research, Boulder, CO}
\address[IOWA]{Dept. of Chemical and Biochemical Engineering, Center of Global and Regional Environmental Research, The University of Iowa, Iowa City, IA}

\address[PennState]{Dept. of Geography and Institute for CyberScience, Geoinformatics and Earth Observation Laboratory, The Pennsylvania State University, University Park, PA}

\begin{abstract}
The \gls{AnEn} method tries to estimate the probability distribution of the future state of the atmosphere with a set of past observations that correspond to the best analogs of a deterministic \gls{NWP}.
This model post-processing method has been successfully used to improve the forecast accuracy for several weather-related applications including air quality, and short-term wind and solar power forecasting, to name a few.
In order to provide a meaningful probabilistic forecast, the \gls{AnEn} method requires storing a historical set of past predictions and observations in memory for a period of at least several months and spanning the seasons relevant for the prediction of interest.
Although the memory and computing costs of the \gls{AnEn} method are less expensive than using a brute-force dynamical ensemble approach, for a large number of stations and large datasets, the amount of memory required for \gls{AnEn} can easily become prohibitive.
Furthermore, in order to find the best analogs associated with a certain prediction produced by a \gls{NWP} model, the current approach requires searching over the entire dataset by applying a certain metric. This approach requires applying the metric over the entire historical dataset, which may take a substantial amount of time.
In this work, we investigate an alternative way to implement the \gls{AnEn} method using deep generative models.
By doing so, a generative model can entirely or partially replace the dataset of pairs of predictions and observations, reducing the amount of memory required to produce the probabilistic forecast by several orders of magnitude.
Furthermore, the generative model can generate a meaningful set of analogs associated with a certain forecast in constant time without performing any search, saving a considerable amount of time even in the presence of huge historical datasets.
\end{abstract}


\end{frontmatter}



\section{Introduction}\label{sec:introduction}

Quick and accurate weather prediction is an essential and critical part for decision-making, in particular when
human lives are at stake. It has been counted that more than 200 people died in 2018 because of all weather and climate events over U.S. and the damages has been estimated to be more than 90 billion dollars \citep{ncdc}. However, these losses would be far higher if scientists had not predicted these events. \gls{NWP} model forecast is usually used as the main tool for weather prediction. However, its utility is limited as it represents only a single plausible future state of the atmosphere. In fact, imperfect initial conditions and model deficiencies can lead to model errors that grow non-linearly during the model evolution. \citet{Lorenz1969} pointed out that, ``the errors in estimating the current state of the atmosphere are due mainly to omission rather than inaccuracy''. In other words, the errors can be related to the gaps in science or computational resources and the uncertainty in input data. However, current NWP models are state-of-the-science models that include the most recent discovered science. So, reducing the effects of uncertain data is a possible option to increase the accuracy and reliability of these models. 

In order to address the forecast uncertainty of deterministic models, integrated probabilistic forecasts has been widely used. As shown by \cite{Hopson14}, the rationale for using a probabilistic forecast is four-fold: 1) the ensemble prediction mean usually has half of the error variance of a single forecast; 2) ensemble forecasts are capable to estimate the likelihood of extreme weather events; 3) ensemble forecasts can be used to generate non-Gaussian forecasts \gls{PDF}; and 4) ensemble spread acts as a representation of forecast uncertainty. These ensemble members can be achieved by changing different parts of the models such as different physical parametrization of the sub-grid physical processes \citep{foley2012current}, dynamic solvers \citep{jablonowski2004adaptive}, and initial conditions \citep{sperati2016application}. For example, different variations of an initial condition can be found by adding perturbations from a stochastic model. Deterministic NWP models realizations with these variations will generate an ensemble of predictions to capture the uncertainty for a specific weather event. Although this Monte Carlo based method is a popular and reliable way of generating probabilistic forecasting, it requires a huge amount of computational resources for solving Navier-Stokes equations for multiple times \citep{schmidt2017practice, syrakos2012numerical, fant2016impact}.

In order to reduce the computational cost of the ensemble forecast generation, another method was suggested by \cite{DelleMonache} which is called Analog Ensemble (AnEn) method. AnEn assumes that the model error for past predictions can be used to correct future predictions. As a result, this technique only requires one single NWP model run and an archive of previous model runs and observed values to correct the current model run (See \ref{sec:AnEn} for more details). AnEn has become a very popular method for correcting forecasts due to its robustness and computational resources efficiency and it has been applied to different projects by National Center for Atmospheric Research (NCAR) in the recent years \citep{Cervone2017}. Although AnEn significantly reduced computational resources, it requires keeping the whole historical datasets of the model and corresponding observed values in the memory. The specific implementation of AnEn is provided by the Parallel Analog Ensemble package \citep{PAnEn2019}. This package has been developed during the course of the past 3 years and successfully applied to temperature forecasts for Eastern United States \citep{hu2019dynamically, balasubramanian2018harnessing}.

However, for newer computer architectures, the amount of memory capacity and memory bandwidth per core is diminishing.
As shown in~\cite{Mahapatra:1999:PBP:357783.331677, Wulf:1995:HMW:216585.216588}, the rate of improvement in microprocessor speed exceeds the rate of improvement in DRAM memory.
This is mostly due to the division of the semiconductor industry into microprocessor and memory camps.
Consequently, microprocessor performance has been improving at a rate of 60 percent per year, whereas the access time to DRAM has been improving at less than 10 percent per year.

As a result, we hypothesized that a similar analog method can be applied using the probability distribution function (PDF) of the historical dataset instead of using the actual data. Indeed, we applied a generative machine learning model to generate the analogous states of atmosphere based on current NWP model forecast. The application of different machine learning methods in Geoscience has been explored in other studies. For instance, Convolutional Neural Network (CNN) models have been used to analyze satellite images (\cite{li2017estimating}) and Recurrent Neural Network (RNN) models for air quality predictions (\cite{deepLearning}, \cite{LSTM}). However, the use of generative machine learning models has been not studied for doing atmospheric probabilistic forecasts. In this study, we apply generative machine learning models for the first time as a new technique of atmospheric probabilistic forecasting. Specifically, probabilistic characteristics of Conditional Variational Autoencoder (CVAE) is utilized in this study to learn the multivariate probability distribution of historical observed dataset based on historical modeled dataset and it can be used to produce ensemble forecasts.  

This paper is organized as follows: Section~\ref{sec:AnEn} summarizes the theory behind the Analog Ensemble method
and its most important practical and computational aspects; Section~\ref{sec:DeepGenerativeModels} introduces the main concepts of deep generatives modelling, mathematical background of Variational Autoencoders and Conditional Variational Autoencoders; Section~\ref{sec:approach} explains how probabilistic forecasting can be performed using Conditional Variational Autoencoders and reports the main technical issues encountered during the process;
Section~\ref{sec:experiment} describes the experimental setting; Section~\ref{sec:results} reports the
results obtained by the experiments and describes the metric used to evaluate and compare the performance of the CVAE against the AnEn; Section~\ref{sec:Discussion} provides more insights about the results and shows the memory footprint and runtime for probabilistic forecasting using CVAE; finally, Section~\ref{sec:conclusion} concludes the paper describing the potential uses of producing probabilistic forecasting with deep generative models and proposing new ideas for future works.

\section{Analog Ensemble}\label{sec:AnEn}

The Analog Ensemble (AnEn) method tries to estimate the probability distribution of the future state of the atmosphere with a set of past observations that correspond to the best analogs of a deterministic \gls{NWP}.
In particular, \gls{AnEn} seek to estimate the probability distribution ~$[f(.)]$ of the observed value of the predictand variable
given a model prediction, which can be represented as in Eq.~\ref{eq:AnEnEq}

\begin{equation} \label{eq:AnEnEq}
  f(y | x ^ f)
\end{equation}

where, at a give time and location, \textit{y} is the observed future value of the predictand variable, and
$ x^f = (x^f_1 , x^f_2 , ...,  x^f_k )$ contains the values of \textit{k} predictors from the deterministic model
prediction at the same location.

The \gls{AnEn} method generates samples of $y$ via three main steps using a history of cases, called the analog training
period, in which both the \gls{NWP} (deterministic forecast) and the verifying observations are available. Step 1 consists in searching and selecting the best-matching historical forecasts for the current prediction, these forecasts are called analogs. An analog may come from any past date within the training period.
Step 2 consists in getting each analog's verifying observation. Step 3 consists in grouping the selected observations, creating the final ensemble prediction for the current forecast.

For Step 1, the quality of an analog is determined by a metric, which is usually represented by the following equation presented in \cite{DelleMonache2011}:

\begin{equation} \label{eq:metric}
    \| F_{t}, A_{t'} \| = \sum_{i=1}^{N_v} \frac{w_i}{\sigma_{f_i}} \sqrt{\sum_{j=-\overline{t}}^{\overline{t}} (F_{i,t+j} - A_{i,t'+j})^2}
\end{equation}
where $F_t$ is the current NWP deterministic forecast valid at the future time t at a station location; $A_{t'}$ is an analog
at the same location and with the same forecast lead time but valid at a past time $t'$; $N_v$ and $w_i$ are the number
of physical variables used in the analogs search and their weights, respectively; $\sigma_{f_i}$ is the standard deviation of the
time series of past forecasts of a given variable at the same location and forecast lead time; $\overline{t}$ is equal to half
the number of additional times over which the metric is computed; and $F_{i,t+j}$ and $A_{i,t'+j}$ are the values of the
forecast and the analog in a time window for a given variable.

During this whole three-step process, the entire historical dataset of past forecasts and observations needs to be kept in memory. Furthermore, the metric expressed in eq.~\ref{eq:metric} used in Step 1 has to be applied to each and every forecast of the historical dataset in order to select the best analogs.
Although the memory and computing costs of the \gls{AnEn} method are less expensive than running a numerical weather prediction model, for a large number of stations and large datasets, the amount of memory required for \gls{AnEn} can easily become prohibitive.

\section{Deep Generative Models}\label{sec:DeepGenerativeModels}

A generative model describes how a dataset is generated, in terms of a probabilistic model. The final goal is to sample from the generative
model in order to generate new data which is consistent with the original model.
A generative model must be probabilistic rather than deterministic. In fact, a generative model should not produce the same output
given the same input. For this reason, a generative model should include a stochastic element that drives the generation of new
samples.

In machine learning, the most common task performed by models is \textit{discrimination}, either in the form of classification
(e.g. is this a picture of a cat or not?) or regression (e.g. what's the prediction for tomorrow's temperature?).
In both cases, a set of features is given as input to the model and the output is either a classification value (label) or a prediction.
In generative models, the model is asked to learn how samples of a certain class look like and to produce new ones based on a
stochastic value.
In a more formal way, discriminative models estimate $p(y | x)$, meaning the probability of having the label $y$ given the input $x$.

Generative models on the other hand, try to estimate $p(x)$: the probability of observing the input $x$.
If the dataset is labeled, a generative model estimates the distribution $P(x|y)$.
This distribution represents the probability of having a set of features $x$ given a label $y$.
In other words, discriminative models attempt to estimate the probability that an observation $x$ belongs to category y.
Generative models do not care about labeling observations. Instead, they try to estimate the probability of seeing the observation at all.
Assuming to be able to build a perfect discriminative model to identify cats vs non-cats, the model would still have no idea how to create a picture (or features like weight and height) that looks like a cat. 
It can only output probabilities against existing images, as this is what it has been trained to do.
We would instead need to train a generative model, which can output sets of pixels that have a high chance of belonging to the original training dataset.

\subsection{Variational Autoencoder}\label{subsec:VAE}

One of the first papers about deep generative models was published by \cite{VAE2013}, which laid the foundations of the well-known deep neural network called \gls{VAE}.
The mathematical basis of Variational Autoencoders has relatively little to do with the traditional
autoencoders. In fact, \gls{VAE} are called "autoencoders" only because of their architecture, composed
by an encoder and decoder, resembles a traditional autoencoder.
In this work, we only provide a high-level view to justify the application of \gls{VAE} for probabilistic forecasts. A very detailed and yet understandable explanation of \gls{VAE} is provided by \cite{TutorialVAE}.

As mentioned in Sec.~\ref{sec:DeepGenerativeModels}, a deep generative model tries to approximate the true
distribution of the inputs using a deep neural network.

\begin{equation}\label{eq:P(x)}
  x \sim P_\theta(x)
\end{equation}

The distribution is expressed in eq. \ref{eq:P(x)}, where $theta$ are the parameters of the distribution determined during training.
In machine learning, it is important to find equation \ref{eq:P(x,z)}, a joint distribution between the inputs
$x$ and the latent variables $Z$. The latent variables are derived by the encoder of the \gls{VAE}
and they represent properties observable from inputs.
This sort of learning is called \textit{representation learning} and it is based on the
idea that each point in the latent space (having less dimensions than the original one) is the representation of some high-dimensional sample, as explained in \cite{Representation_lrn}.
In other words, eq. \ref{eq:P(x,z)} is a distribution of input data points and their attributes.

\begin{equation}\label{eq:P(x,z)}
  P_\theta(x,z)
\end{equation}

The main distribution in eq. \ref{eq:P(x)} can be computed as expressed in eq. \ref{eq:P(x)_integral}

\begin{equation}\label{eq:P(x)_integral}
  P_\theta(x) = \int{P_\theta(x,z) dz}
\end{equation}
which means considering all of the possible attributes in order to describe the input.
This problem is not tractable, because eq. \ref{eq:P(x)_integral} cannot be solved analytically.
By using the Bayes theorem, equation \ref{eq:P(z|x)_integral} represents an alternative expression for eq. \ref{eq:P(x)_integral}:

\begin{equation}\label{eq:P(z|x)_integral}
  P_\theta(x) = \int{P_\theta(z|x) P(z) dz}
\end{equation}

The goal of the \gls{VAE} is to find a tractable distribution that estimates $P_\theta (z|x)$.

In order to make $P(z|x)$ tractable, \gls{VAE} makes use of an "encoder" able to generate
the distribution $Q_\phi$ defined in the following equation:

\begin{equation}\label{eq:Q(z|x)}
  Q_\phi(z|x) \approx P_\theta(z|x)
\end{equation}

$Q$ can be approximated by a neural network and it is chosen to be a multivariate Gaussian, as
expressed in the following equation:

\begin{equation}\label{eq:Q(z|x)_normal}
  Q_\phi(z|x) = \mathcal{N}(z;\mu(x),diag(\sigma(x) ))
\end{equation}

where $\mu$ and $\sigma$ are computed by the encoder using the input dataset.

$Q_\phi(z|x)$ generates the latent vector z from the input x and it represents the encoder of the
\gls{VAE}. On the other hand, $P_\theta(x|z)$ generates the input from the latent vector z and thus
it acts as decoder.
Our final goal is still to estimate $P_\theta(x)$.

In order to quantify the distance between $ Q_\phi(z|x)$ and $P_\theta(z|x)$, the
\gls{KL} divergence is used.
\gls{KL} determines the distance between our two conditional densities, as expressed in the
following equation:

\begin{equation}\label{eq:KL}
  D_{KL}(Q_\phi(z|x) \| P_\theta(z|x)) = \mathbb{E}_{z \sim Q}[\log Q_\phi(z|x) - \log P_\theta(z|x)]
\end{equation}

By using the Bayes theorem and rearranging a few terms, equation \ref{eq:KL} becomes:

\begin{equation}\label{eq:KL_core}
  \log P_\theta(x) - D_{KL}(Q_\phi(z|x) \| P_\theta(z|x)) = \mathbb{E}_{z \sim Q}[\log P_\theta(x|z)] - D_{KL}(Q_\phi(z|x) \| P_\theta(z))
\end{equation}

Equation \ref{eq:KL_core} represents the core equation of the \gls{VAE}.
On the left-hand side, the term $P_\theta(x)$ is the term we are trying to estimate minus the error imposed
by $Q_\phi(z|x)$: the approximation of the real $P_\theta(z|x)$. 
When the approximation of $P_\theta(z|x)$ is good, the KL distance goes to zero. This first part of the equation
represents the encoder of the \gls{VAE}.
On the right-hand side, $P(x|z)$ represents the decoder of the \gls{VAE}, whereas the DL distance represents the loss
function from the Gaussian distribution $Q_\phi(z|x)$ expressed in eq. \ref{eq:Q(z|x)_normal}.

The left-hand side of eq. \ref{eq:KL_core} is also known as \gls{ELBO}, and because the KL divergence is always positive,
it represents a lower bound for $log P_\theta(x)$. When the encoder and decoder of the VAE are trained together, the \gls{ELBO}
is maximized; meaning that the KL distance on the left-hand side of the equation goes to zero (thus the encoding of x in z is
getting better) and that $log P_\theta(x|z)$ on the right-hand side is maximized
(thus the decoder is getting better in reconstructing x from the latent representation z).

The right-hand side of eq. \ref{eq:KL_core} has two important parts of the final loss function of the \gls{VAE}:
the decoder part $\mathbb{E}_{z \sim Q}[\log P_\theta(x|z)]$ takes samples from the output of the encoder to reconstruct
the inputs. Maximizing this term means minimizing the reconstruction loss $\mathcal{L}_R$.
The second part $- D_{KL}(Q_\phi(z|x) \| P_\theta(z))$ is straightforward to evaluate, thanks to the Gaussian nature
of $Q_\phi(z|x)$, becoming:

\begin{equation}\label{eq:KL_core_gauss}
  - D_{KL}(Q_\phi(z|x) \| P_\theta(z)) = 0.5 \sum_{j=1}^{J} (1 + \log(\sigma_j)^2 - (\mu_j)^2 - (\sigma_j)^2) 
\end{equation}

where $J$ is the dimension of $z$. This quantity is called KL Loss, represented as $\mathcal{L}_{KL}$.

Summarizing, the loss function of the \gls{VAE} is $\mathcal{L}_{VAE} = \mathcal{L}_{R} + \mathcal{L}_{KL}$.

Although theoretically solid, in practice this approach usually leads to a degenerate solution where
$ Q_\phi(z|x) \approx P_\theta(z) $; meaning that the variational posterior does not depend on the data (x and z are basically independent).
In other words, the model does not learn a good representation of the data.
This problem is known as KL-vanishing and it will be covered in Sec.~\ref{subsubsec:klvanishing}.

An alternative way to see the \gls{ELBO} is as a regularized version of the regular autoencoder.
In particular, $\mathcal{L}_{R}$ represents the reconstruction loss of the regular autoencoder and $ \mathcal{L}_{KL} $
the regularization.
From this point of view, it is natural to introduce a new hyperparameter $\beta$ able to control the strength of the regularization,
leading to the following equation:

\begin{equation}\label{eq:beta-vae}
   \mathcal{L}_{VAE} = \mathcal{L}_{R} + \beta \mathcal{L}_{KL}
\end{equation}

By changing the value of $\beta$ from 0 to 1 (or greater than 1) we can nullify (transforming the VAE in an AE) or force more the posterior on the latent representation.
As explained by \cite{cyclic-kl}, by cyclically assigning values between 0 and 1 to $ \beta $ it is possible to mitigate the KL-vanishing problem.

When $\beta > 1$ the VAE is referred as $\beta$-VAE by~\cite{Higgins2017betaVAELB}.
As explained in~\cite{disentangling-factorising} and~\cite{isolating-disentanglement}, a value of $\beta > 1$ can be used to force the VAE
to learn disentangled representations of the data in the latent representation z.

\subsection{Conditional Variational Autoencoder}\label{subsec:CVAE}

In a \gls{VAE}, if the latent space is randomly sampled, the \gls{VAE} has no control over which the kind of data
to generate.
For example, in a \gls{VAE} trained to generate hand written digits over the MNIST dataset, there is no control
over which digit should be produced.

The approach proposed in \cite{Deep_CVAE}, called \gls{CVAE}, solves this problem by imposing a condition on
both encoder and decoder inputs.
Equation \ref{eq:KL_core} is modified to include the condition in each term as follows:

\begin{multline}
\label{eq:KL_core_contitional}
  \log P_\theta(x|c) - D_{KL}(Q_\phi(z|x,c) \| P_\theta(z|x,c)) \\
  = \mathbb{E}_{z \sim Q}[\log P_\theta(x|z,c)] - D_{KL}(Q_\phi(z|x,c) \| P_\theta(z|c))
\end{multline}

Similar to a classic \gls{VAE}, a \gls{CVAE} tries to estimate the quantity expressed in the following equation:

\begin{equation}\label{eq:P(x)_contitional}
  P_\theta(x|c)
\end{equation}

which would allow us to generate samples from the distribution of the input data under a certain condition.

\section{Proposed Approach}\label{sec:approach}

The main focus of this work is to generate probabilistic forecasting without maintaining the entire dataset of forecasts and
observations in memory. Generative models are a great fit for learning the distribution of the original dataset and generating
new data.
A valid way to generate probabilistic forecasting using deep generative models is to transform the \gls{AnEn} method
from an instance-based (~\cite{instance-based-learning}) to a generative-based method.

Equation \ref{sec:AnEn} represents the probability distribution of the observed value of the predictand variable $y$ given
a model prediction $x$, composed by various predictors. This equation can be easily seen as equation \ref{eq:P(x)_contitional}
of a \gls{CVAE}, where the condition $c$ is the forecast coming from the deterministic \gls{NWP} and $x$ is the observation
coming from the input data.

For a correctly and successfully trained \gls{CVAE}, we can perform probabilistic forecasting without keeping in memory
the whole dataset and without performing any search of the best-matching analogs.
This is because the \gls{CVAE} has been trained to generate samples that already match as best as possible the distribution
of the input data, conditioned to the forecast provided as condition.

As mentioned in Sec.~\ref{sec:introduction}, the AnEn method is very popular for correcting forecasts produced by a numerical weather model.
In this work, we focus on correcting only the wind speed by using four predictors: wind speed, wind direction (expressed in radians), 2 meter temperature and pressure.

\subsection{CVAE Training}\label{subsec:CVAE-training}

Given a dataset $\mathcal{D}$, composed by observations $o$ paired with forecasts $f$ produced by a \gls{NWP} model,
the training phase for a \gls{CVAE} can be performed by passing the observations as the data to be generated by the model,
and the forecasts as the conditions. In formulas, $P_\theta(x|c)$ becomes $P_\theta(o|f)$.

\begin{figure}[h!]
\begin{center}
   \includegraphics[width=0.8\linewidth]{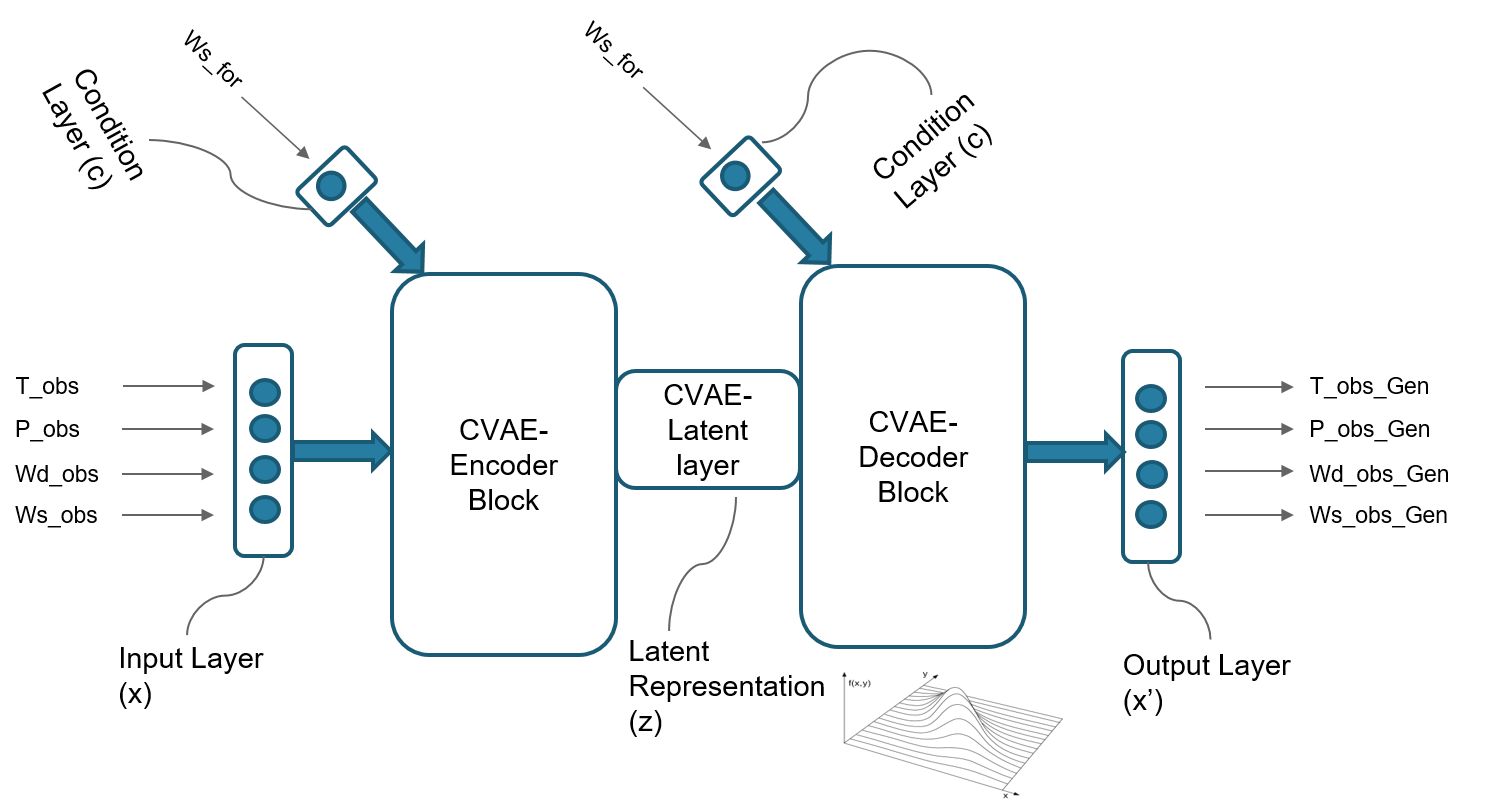}
\end{center}
   \caption{CVAE general architecture - training}
\label{fig:model_trainedCVAE}
\end{figure}

Fig.~\ref{fig:model_trainedCVAE} shows the structure of the CVAE during training: the variables that we want to generate,
coming from the observations, are the inputs of the CVAE (e.g. T\_obs represents the observed temperature).
The condition is the forecast produced by the \gls{NWP} model, associated with the observations given as input (e.g. Ws\_for stands for
Wind Speed forecast). The same condition is given to the decoder during the training phase, this allows the decoder to tie a specific condition
with a certain latent representation).
The output of the decoder layer is the reconstructed features given in input.

\subsubsection{KL-Vanishing Problem and Cyclic Annealing Scheduling}\label{subsubsec:klvanishing}

Training a CVAE with traditional methods can be challenging, in particular because of a notorious issue already experienced by~\cite{Bowman2015,Yang2017}: the decoder ignores the latent variable, yielding what is termed the KL-vanishing or latent variable collapse problem (~\cite{avoid-latent-variable-collapse}).
As mentioned in Sec.~\ref{subsec:VAE}, it is possible to mitigate the KL-vanishing problem by cyclically changing the value of $\beta$ during the training
as explained in details by~\cite{cyclic-kl}.

The KL-vanishing problem is supposed to be related to the low quality of z at the beginning of the decoder training.
As a result, the model is forced to learn an easier solution by relying only on the previous samples of x without relying on z at all.

During the training of our CVAE with $\beta$ fixed to 1, we observed the KL-vanish problem since the very first epochs.
As mentioned by~\cite{how-to-train-vae} and~\cite{Bowman2015}, setting $\beta = 0$ for the first epoch and then fix it to 1 for the rest of the epochs made
a huge difference in the final results. In fact, this approach called Warm-up transforms the CVAE in a regular autoencoder for the first epoch,
creating a more meaningful z than random values to be used by the final CVAE when $\beta$ is set to 1 (monotonic annealing schedule).
The cyclic annealing scheduling proposed by~\cite{cyclic-kl} produced even better results for our CVAE.
This approach proposes to cyclic several times between $\beta = 0$ and $\beta = 1$ during the epochs.
By doing so, the values of z is very close to $Q_{\theta}(z | x)$ at the beginning of the training, which is very meaningful.

\begin{figure}[h!]
\begin{center}
   \includegraphics[width=1.0\linewidth]{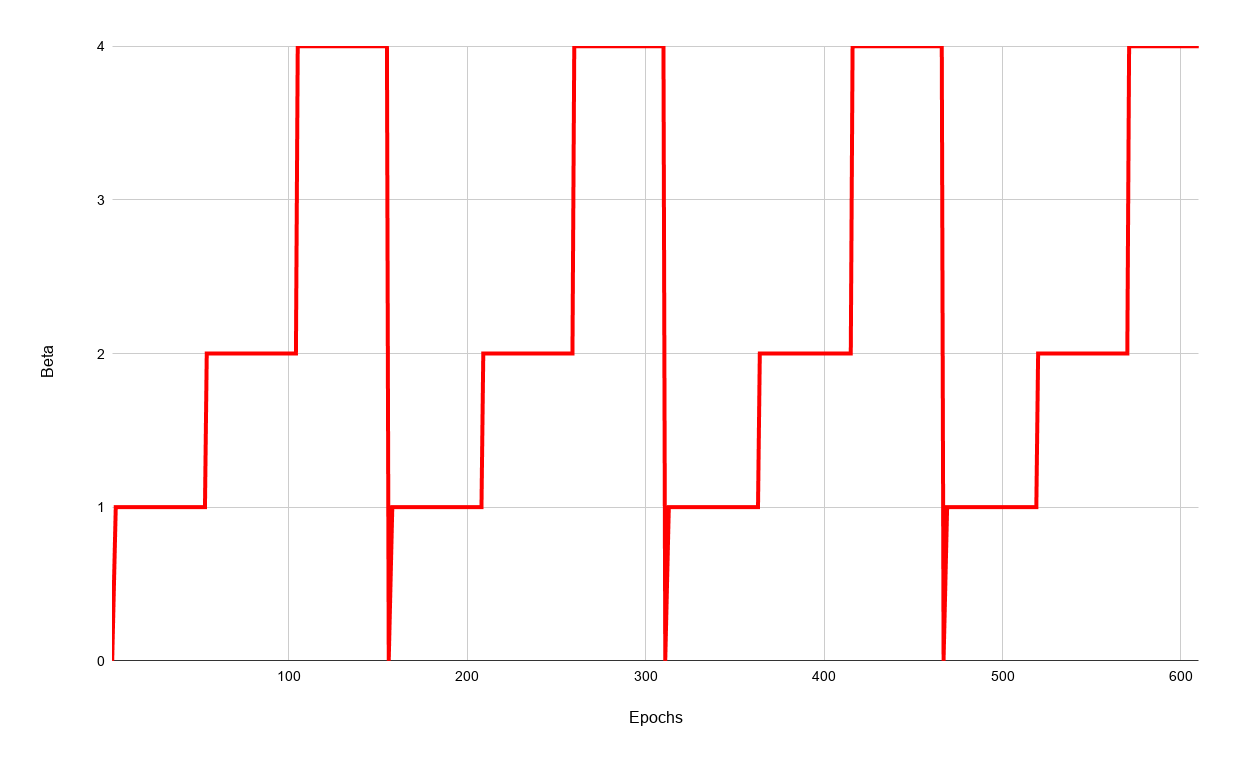}
\end{center}
   \caption{Cyclic Annealing with Beta $>$ 1}
\label{fig:beta_vs_epochs}
\end{figure}

For our CVAE, we obtained the best results by using a cyclic annealing scheduling including values of $\beta > 1$.
In particular, we adopted the cyclic scheme shown in Figure~\ref{fig:beta_vs_epochs} where only one epoch was executed by setting $\beta = 0.0$ and 0.5; for $\beta = 1.0$, 2.0 and 4.0 we executed 50 epochs in each case. 

\subsection{CVAE Inference}\label{subsec:CVAE-inference}

Inference or sample generation in VAE and CVAE is performed only by the decoder.
In order to generate a new data point, z-dimension random numbers $r$ sampled by a Normal distribution with
mean 0 and variance 1 represent the vector z in the latent space.
This vector of random numbers is given as input to the decoder along with the forecast as condition of the CVAE.

\begin{figure}[h]
\begin{center}
   \includegraphics[width=0.8\linewidth]{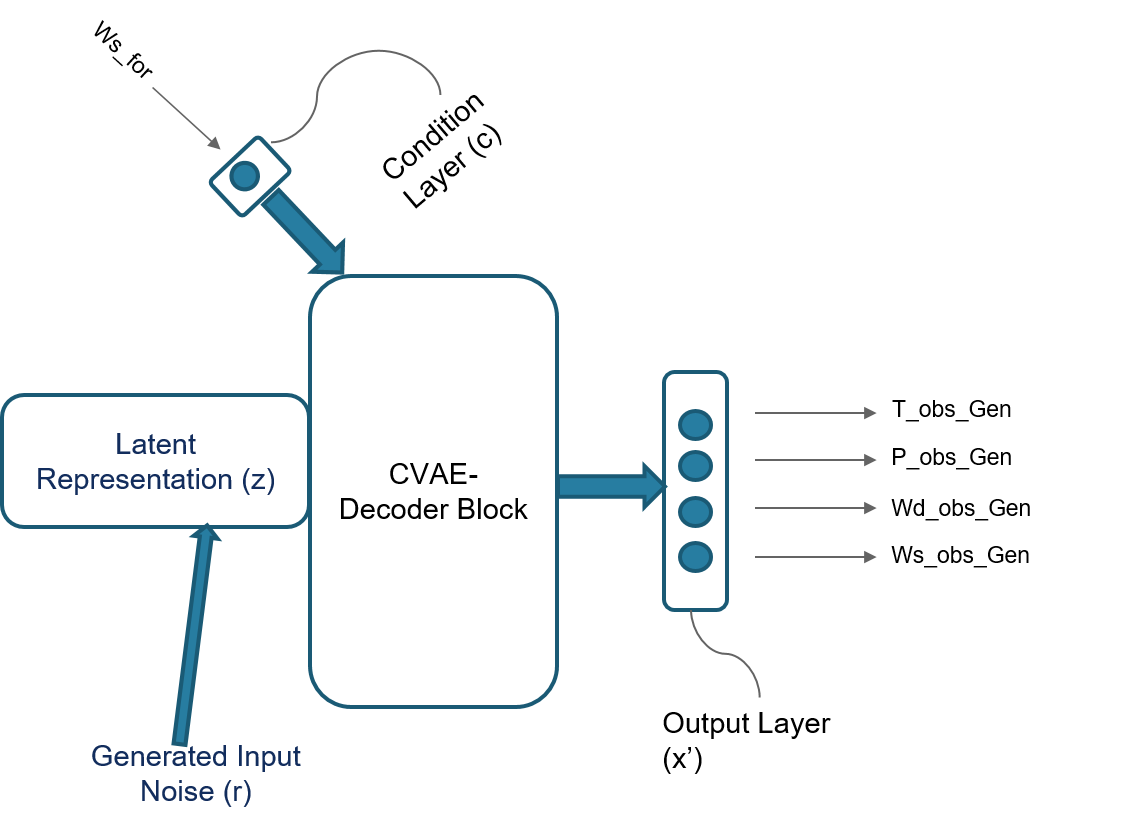}
\end{center}
   \caption{CVAE decoder architecture - inference}
\label{fig:model_testedCVAE}

\end{figure}

Figure~\ref{fig:model_testedCVAE} shows the decoder architecture able to generate analogs according to the condition "Ws\_for".

This simple two-step method allows us to make probabilistic forecast in constant time and memory.

\section{Experiments}\label{sec:experiment}

As the case study in this work, we seek to correct the forecasted wind speed value using observational dataset. As a result, we feed NWP forecasted wind speed to the model as the condition layer. \cite{DelleMonache2013} found that 10-m wind speed (Ws) and direction (Wd), 2-m temperature (T), and surface pressure (P) were the most important parameters in AnEn for 10-m wind speed prediction. As a result, the same variables have been used in this study for input and output layers.

A 10*5 box in NAM dataset covering State College, PA, is chosen for running all the tests. As forecast dataset, we used data from The North American Mesoscale Forecast System (NAM) (\cite{NAM}) which is one of the major operational forecasting systems maintained by NOAA(\cite{noaa}). NAM initiates every day at 00 UTC and predicts meteorological status of the next 84 hours. The outputs are saved in hourly format for the first 36 hours and 3-hourly for the next 48 hours; in other words, 53 leading times. On the other hand, NOAA uses real observed data such as satellite, aircraft, and ground measurements data and by assimilating them into NAM forecasts, it provides another dataset called NAM Analysis (\cite{noaa}). NAM Analysis usually has data for 16 hours per day and we used this dataset as the observation dataset in this study. 

Although longer periods have been suggested for probabilistic forecasts (\cite{DelleMonache2013}, our analysis shows that NAM model configurations have been updated roughly every year. Moreover, consistent data are required for training machine learning models. As a result for training the model, NAM forecasts and analysis data for 365 days between 2017-06-01 and 2018-06-01 are used for training. A 7-days validation period between 2018-06-01 and 2018-06-08 has been used for tuning and analyzing the probabilistic performance of the model. Different configurations in terms of number of layers and number of neurons has been tested and a configuration with reasonable size and performance has been selected as the proposed model (CVAE). Moreover, the performance of the proposed model with bigger datasets has been evaluated and discussed in discussion section. 

\section{Results}\label{sec:results}

The focus of this paper is on generating probabilistic prediction of wind speed. As a result, the following evaluations are based on the generated (for CVAE) and analogs (for AnEn) values of wind speed. The performance of the model for other output variables is discussed in Section \ref{sec:Discussion}. In order to evaluate the performance of the ensemble forecasts generated by the CVAE and compare the results with the AnEn ensemble analogs, we have used the metrics for reliability and probability scores in ensemble forecasting. Each of the 7 days composing the testset has predictions for the next 84 hours; the following results are based on the average of these 7 days for each leading time. It should be mentioned that the observations data (NAM analysis) do not change for the first 4 leading times. This can be confusing for the model and also do not provide meaningful information and thus, we have excluded them from the dataset for both CVAE and AnEn. The CVAE trained model uses Gaussian noise in order to produce one forecast as explained in \ref{subsec:CVAE-inference}. However, 21-members ensemble forecast need 21 forecasts. As a result, the CVAE has to be used for inference 21 times per each forecast, using new generated noise sampled from a normal distribution (N(0,1)).

\subsection{Dispersion and Rank Histogram}\label{subsec:rankHistogram}

In probabilistic forecasting, a reliable forecast seeks to have the same relative frequency as the probability of the forecasted value (\cite{DelleMonache2013}). This means that the reliability can be expressed as the conditional probability of the observed value given that the forecasted value actually happens (\cite{Jolliffe}):

\begin{equation}\label{eq:reliability}
  Pr[Y = k |F = p] = P_k
\end{equation}

In eq. \ref{eq:reliability}, Y and k are the observation random variable and observed value, F and p are the forecast random variable and predicted value. In ensemble probabilistic forecasts, two criteria are usually studied in terms of reliability (\cite{Sperati2017}; \cite{Cervone2017}; \cite{ALESSANDRINI2015}): Rank Histogram (RH) and Dispersion. 

\begin{figure}[htb!]
\centering
   \includegraphics[width=0.8\linewidth]{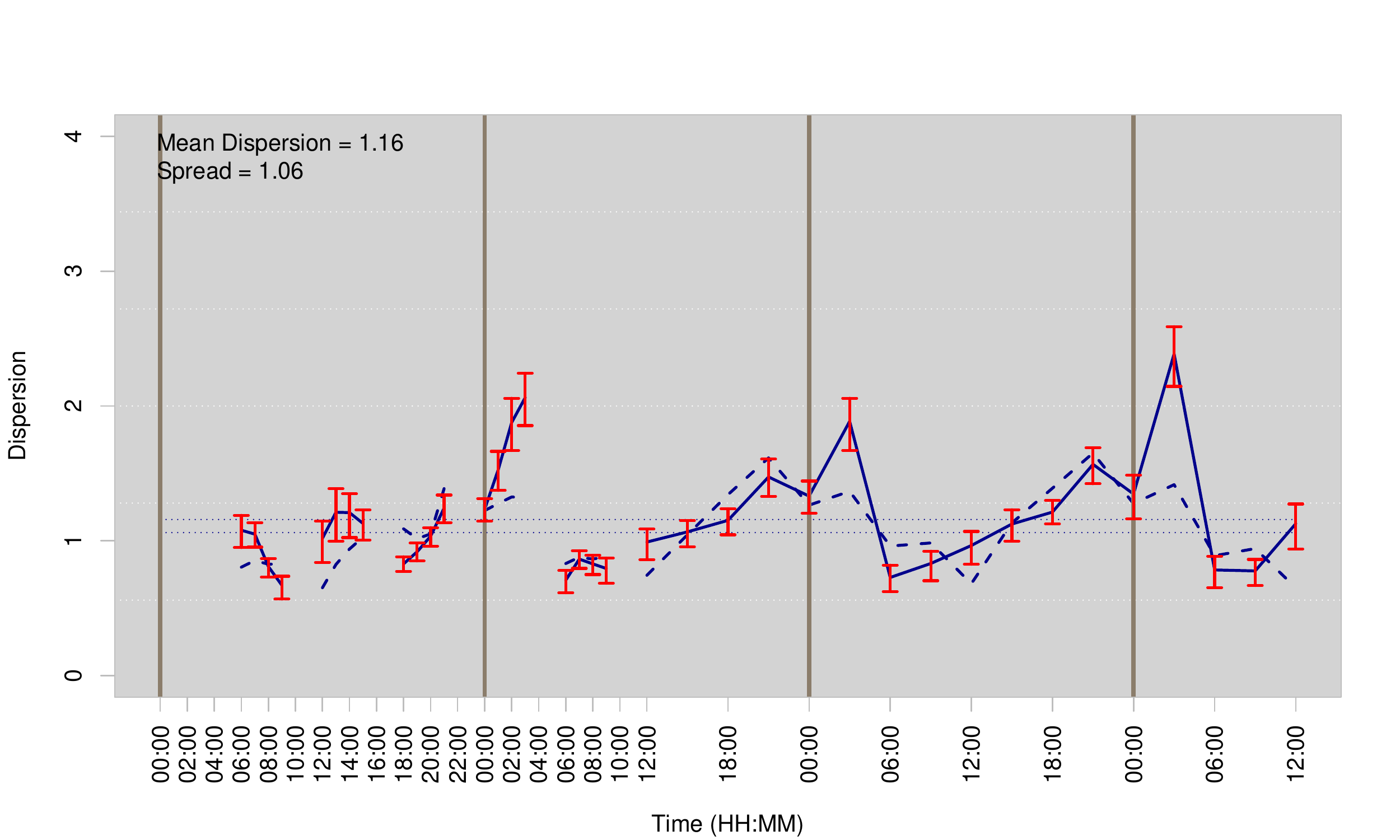}
   \caption{Ensemble members Mean Squared Error (MSE) (solid line) with bootstraping (red range) and mean ensemble variance (dashed line) for a) CVAE, b) AnEn}
   \label{fig:dispersionAnEn}
\label{fig:dispersion}
\end{figure}

Dispersion is one of the simplest ways to get an insight of the reliability of an m-member ensemble forecast (\cite{Jolliffe} and references therein). Dispersion relates that Mean Squared Error (MSE) and mean ensemble variance should be close to each other (with the ratio of m+1/m) in a reliable forecast. In this study, a bootstrap method with 1000 samples has been applied for mean dispersion and CRPS (see section \ref{subsec:CRPS}) in order to consider uncertainty. Figure \ref{fig:dispersion} depicts 50-stations averaged MSE (mean dispersion) for each leading time averaged during testing period for CVAE and AnEn. AnEn has lower all-leading-times averaged MSE which suggests its better prediction performance compared to CVAE. On the other hand, CVAE Mean dispersion oscillates more in different leading times whereas the mean ensemble variance (spread) does not while AnEn dispersion plot shows that MSE and spread have similar trends during leading times. However, the difference between mean dispersion and spread for both CVAE (0.05) and AnEn (0.10) are very close to each other. This suggests that both models are providing similar reliable probabilistic characteristics in terms of dispersion. 

\begin{figure}[htb!]
\centering
   \includegraphics[width=0.8\linewidth]{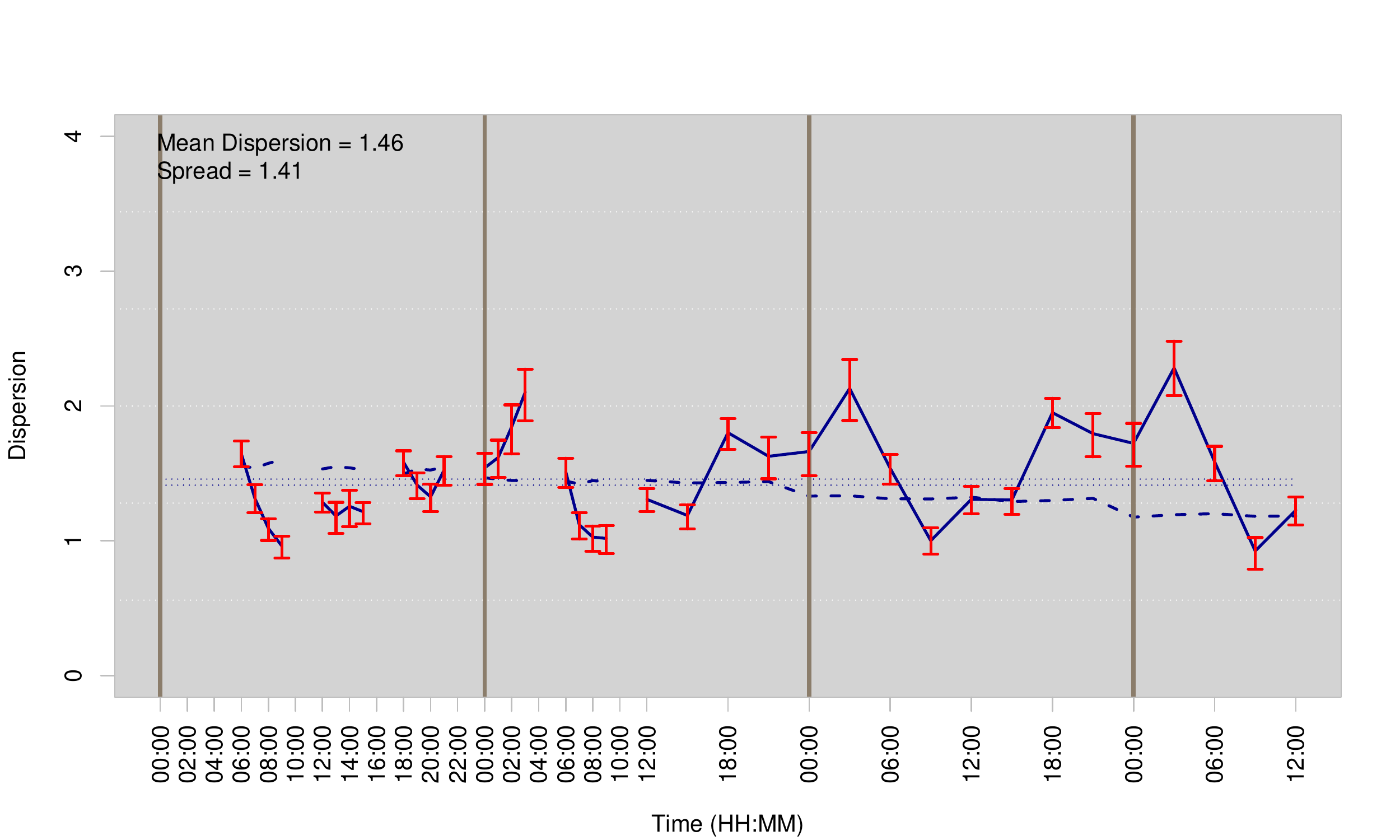}
   \label{fig:dispersionCVAE} 
   \end{figure}
   
In order to get further insights on the reliability of two models, RH has been studied, too. In order to calculate RH, an observed value is ranked based on its corresponding ensemble members and the histogram after giving ranks to all the observed values is plotted (\cite{Jolliffe}). The processes of plotting RH is as follows: in this study, we have 21 predicted values (21-members ensemble) and one observed actual value per forecast; for a total of 22 values. These values are sorted in ascending order and the rank of the observed value is found. After doing the same task for all the forecasts, the histogram of ranks is plotted. For a reliable probabilistic ensemble forecast with m members, the ensemble members will be statistically indistinguishable with a constant ratio of 1/m+1 (0.045 in this study) (\cite{Jolliffe}). 

Figure \ref{fig:others} shows the RH for AnEn and CVAE for the 7-days testing period considering all the stations. These plots show that although both models are having flat frequencies in some portions of the histograms, neither CVAE nor AnEn are perfect reliable forecasts. Specifically, RH results for AnEn show that there is a skew of the predictions towards lower than the observed (real) value. On the other hand, right-skew of CVAE results indicate over prediction of ensemble members. However, roughly distributed RH shows the probabilistic performance of CVAE model. Overall, the range of the frequencies suggests that both models are reliable.

It should be mentioned that \cite{DelleMonache2013} and \cite{Sperati2017} found very uniform RH for wind speed prediction using AnEn which can be related to the size of the historical dataset (\cite{DelleMonache2013}). However, for this study only one year of data was used over a period where the NAM model was not changed. Since the CVAE does not use the actual data and instead uses the characteristics of the data for forecasting, it is crucial that used data during learning process have been produced by one single model configuration. The AnEn method is not very sensitive to these configuration modifications thanks to its instance-based nature, but this is not true for the CVAE with current configuration.
In fact, by using forecasts produced by different versions of NAM, the CVAE is exposed to different distributions during the training.

\begin{figure}[htb!]
\centering
   \includegraphics[width=0.8\linewidth]{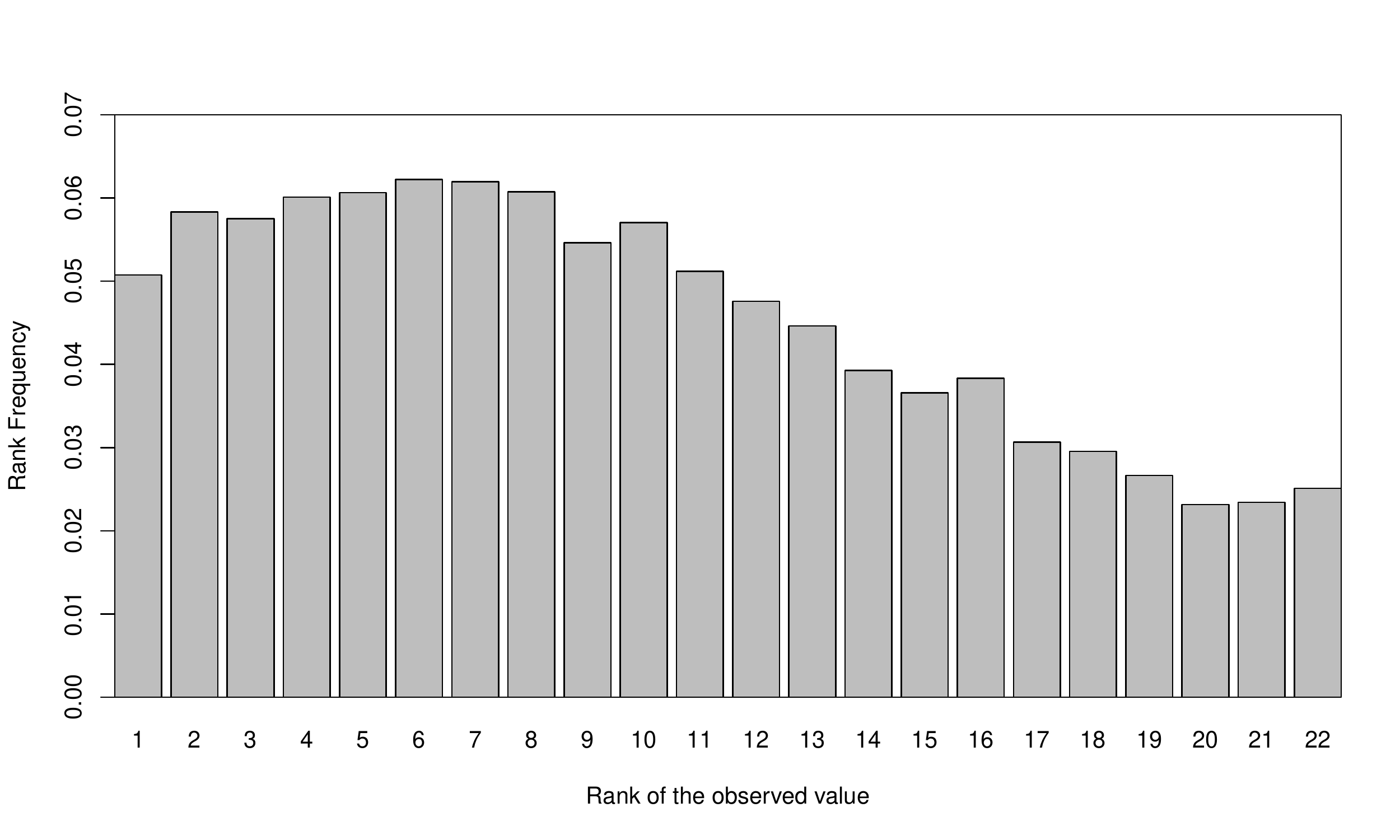}
   \label{fig:histogramCVAE} 
\end{figure}
\begin{figure}[htb!]
\centering
   \includegraphics[width=0.8\linewidth]{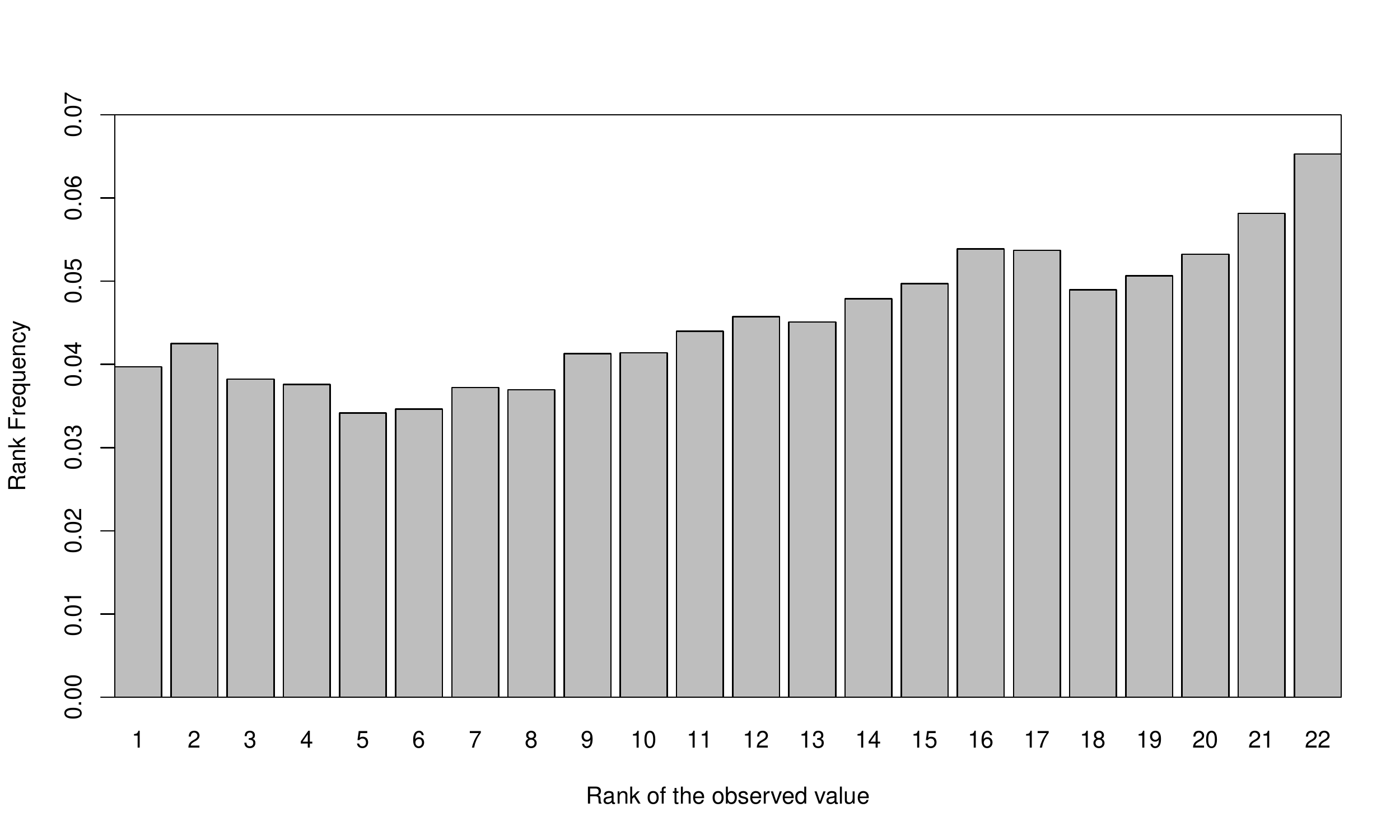}
   \caption{21 ensemble members Rank Histogram for a) CVAE, b) AnEn}
   \label{fig:histogramAnEn}
\label{fig:histogram}
\end{figure}

\subsection{Probability Score}\label{subsec:CRPS}

A number of probabilistic metrics are used to evaluate the performance of a forecast using scores (\cite{Wilks}). Verification statistics are designed to assess the reliability and skill of a forecast (\cite{Jolliffe}). The  Continuous Ranked Probability Score (CRPS) compares the cumulative distribution function (CDF) of the forecast and observation for all the possible outcome values (\cite{Jolliffe}.

\begin{equation}\label{eq:CRPS}
  S(F,o) = \int{(F(z) - H(z-o))^2 dz} 
\end{equation}

\begin{figure}[htb!]
\begin{center}
   \includegraphics[width=0.8\linewidth]{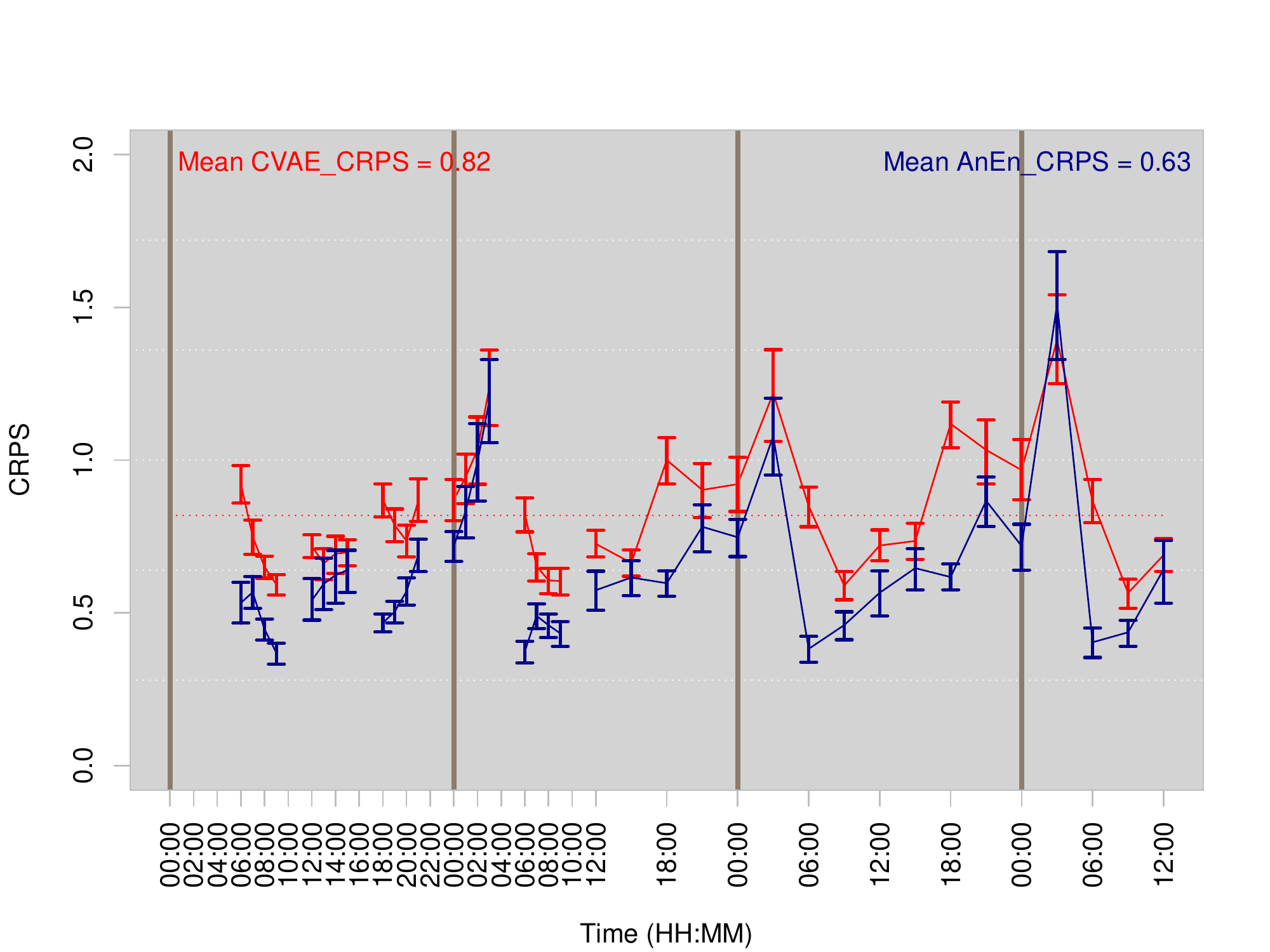}
\end{center}
   \caption{CRPS results for CVAE 21 ensemble members}
\label{fig:CRPS}
\end{figure}

In eq.~\ref{eq:CRPS}, S represents the CRPS scoring rule, F is the cumulative distribution function, H(a), the Heaviside function, is 1 if a $<=$ 0 and zero otherwise, and o is the observed value. However, for obtaining the averaged CRPS, the expectation of the above equation has to be calculated using the concept of decomposition which is explained in detail in \cite{Jolliffe} and references therein. Figure \ref{fig:CRPS} shows CRPS for CVAE and AnEn at each leading time with bootstrapping. CRPS results are lower for absolute values of AnEn at almost all the leading times compared to CVAE. However, both models are showing similar trends. Specifically, the confidence intervals of both models overlap in some of the lead times, but also do not in others. When the confidence intervals overlap, we can assume that the models are statistically similar.

\section{Discussion}\label{sec:Discussion}

The probabilistic verification metrics of CVAE suggest that consistent and reliable predictions of wind speed can be made using the proposed method. While these metrics alone capture only some of the characteristics of good probabilistic predictions, they suggest that the method proposed can be used to generate reliable analogs.

Although the AnEn displays better performance, the trade-off is the size of the data required and the time required for making the forecasts. Recall that CVAE generates ensemble members using the compact representation learned from the datasets, and thus is much more efficient in computational terms.

\begin{figure}[htb!]
\begin{center}
   \includegraphics[width=0.8\linewidth]{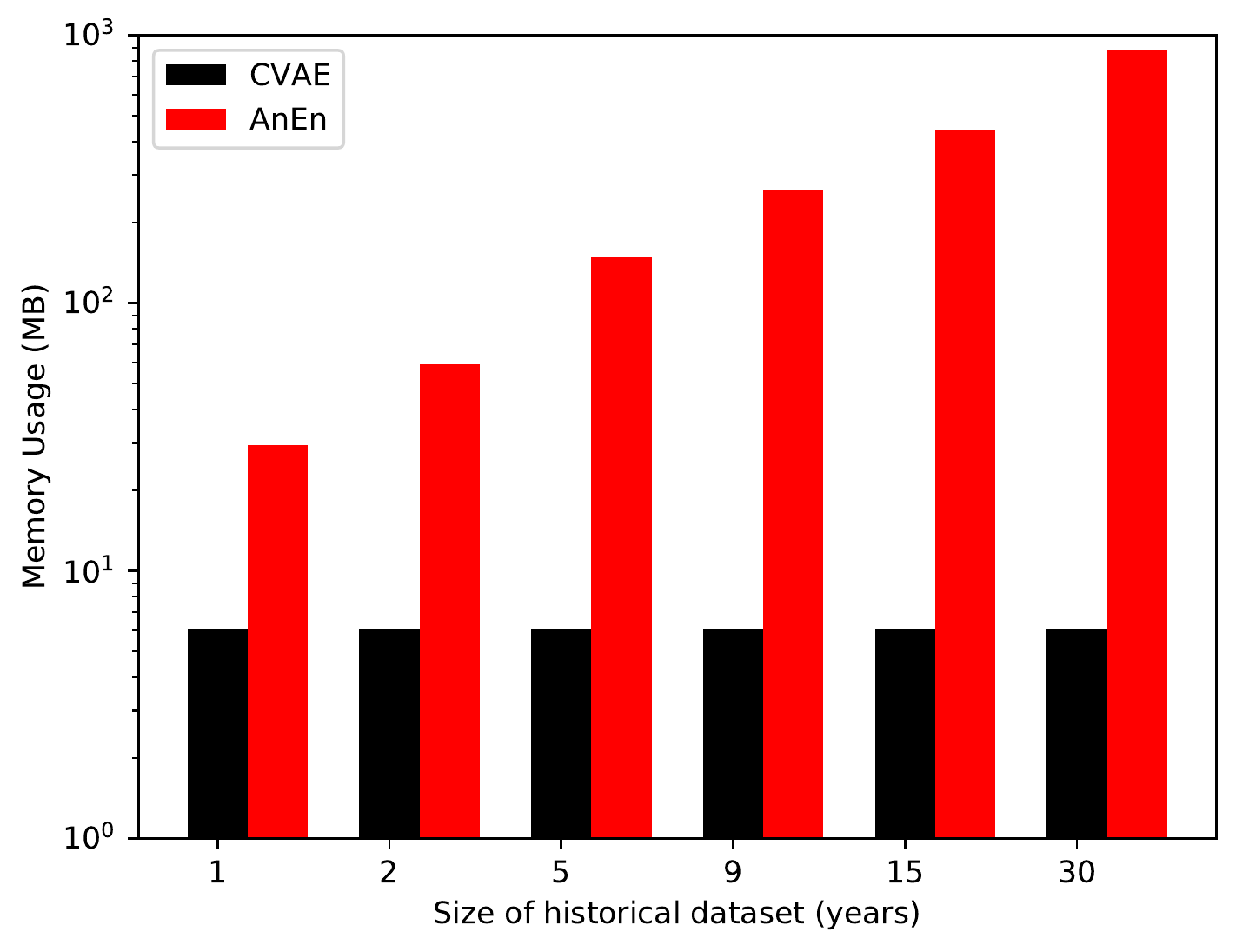}
\end{center}
   \caption{Memory usage comparisons between CVAE and AnEn models}
\label{fig:memory}
\end{figure}

\begin{figure}[htb!]
\begin{center}
   \includegraphics[width=0.8\linewidth]{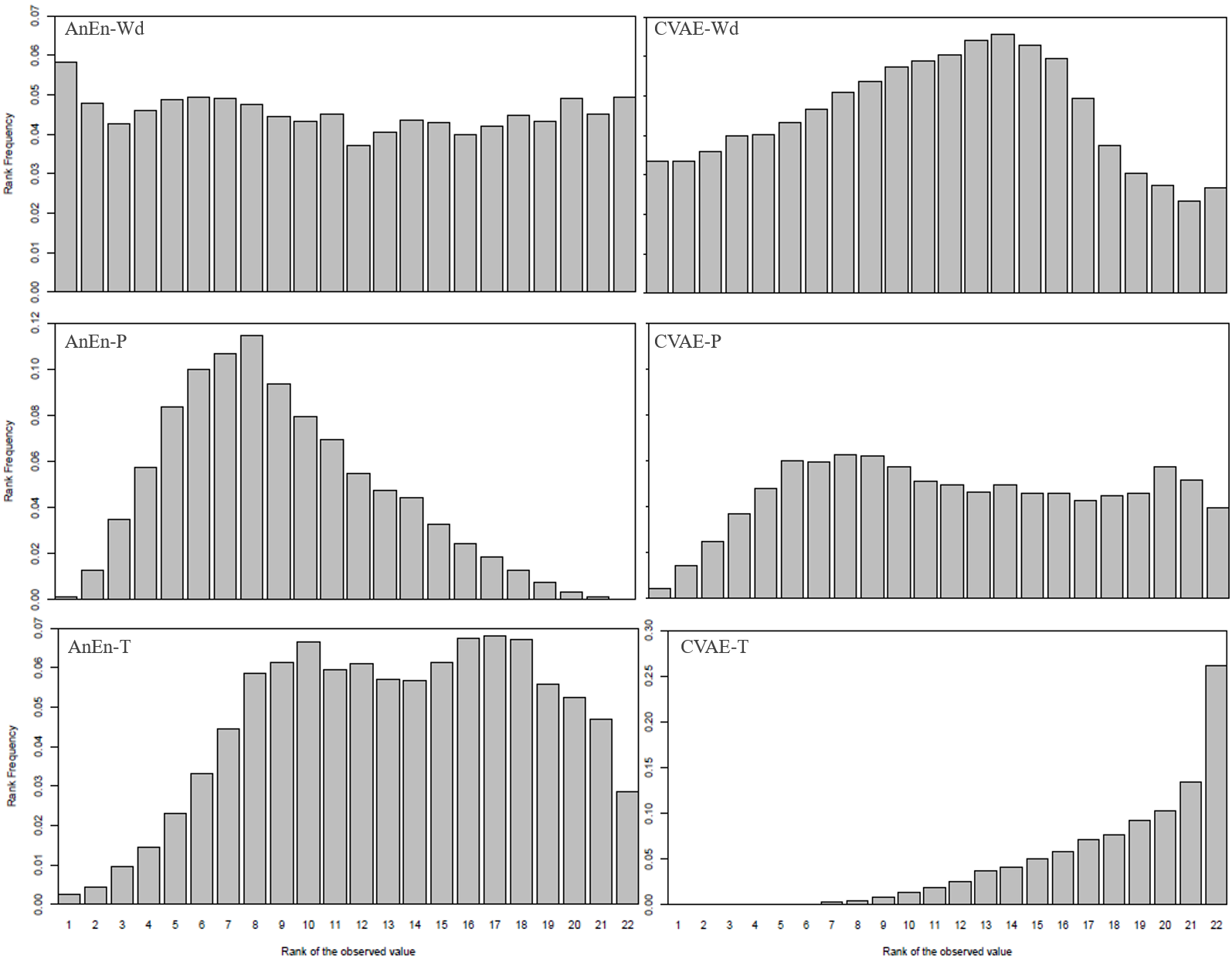}
\end{center}
   \caption{RH for wind direction (Wd), surface pressure (P), and 2-m temperature (T) produced by AnEn (left column) and CVAE (right column) models}
\label{fig:others}
\end{figure}

As discussed in Section~\ref{sec:approach}, we used wind speed as the condition for the CVAE and generated ensembles for 2-m temperature, surface pressure, wind direction, and wind speed in the decoder block. The performance of the proposed approach for probabilistic forecasting of wind speed is discussed in detail in Section~\ref{sec:results}.
In order to provide a more comprehensive evaluation of the observations generated by the CVAE, in Figure~\ref{fig:others} we shows the RH
for wind direction, pressure and 2-m temperature for CVAE and AnEn.

AnEn produces more reliable wind direction values compared to CVAE, whereas CVAE has better performance for surface pressure. Moreover, the results for 2m temperature show that the CVAE is unable to capture the probabilistic behavior whereas AnEn has better performance. The reason for relatively poor performance of the CVAE for variables other than the desired variable can be related to the design of the model. In other words, only the desired variable (wind speed) is feed to the model as the condition and the model optimizes the results only for that variable.
During our tests we observed that providing the whole set of forecast variables as condition during training and inference confuses the model and produces worse results than just providing the goal variable alone (wind speed). 

However, the main motivation for this paper was to show the efficiency of generative machine learning models in probabilistic forecasts in terms of computations and memory footprint. Since the results confirms the probabilistic performance of CVAE, we studied its efficiency in using computational resources. Specifically, memory usage and predictions runtime has been analyzed.

Another advantage of the proposed CVAE solution is that the model needs to be trained only once.
As more data becomes available, incremental training is still possible (encoder block must be saved to retrain CVAE) and useful; in fact, new cases (in particular if extreme cases) can dramatically change the data distribution and final prediction.

Moreover, the training of the model is an offline process. In other words, once the model is trained, it can be used for predictions without keeping the original data. As a result, the memory that will be used in the prediction phase is the size of the model architecture and its weights. Hence, as shown in fig. \ref{fig:memory}, the memory usage for using CVAE does not depend on the size of the historical dataset. 

The size of the model will not change since the architecture (number of layers and neurons) of the model remains constant. In contrast, the memory usage for AnEn depends on the amount of the data to be kept in memory. Specifically, AnEn requires two datasets (pairs of historical forecasts and observations) to be able to find the desired ensemble members; in the case of 1 year the memory needed is about 30 MB and it can reach up to about 900 MB for 30 years. On the other hand, the size of the CVAE models for all 50 stations used in this study is less than 7 MB and it does not change regardless of the size of the historical dataset used to train the model. It should be mentioned that the size of the NAM testing data has not been considered since both models have to keep it in memory.

\begin{figure}[htb!]
\begin{center}
   \includegraphics[width=0.8\linewidth]{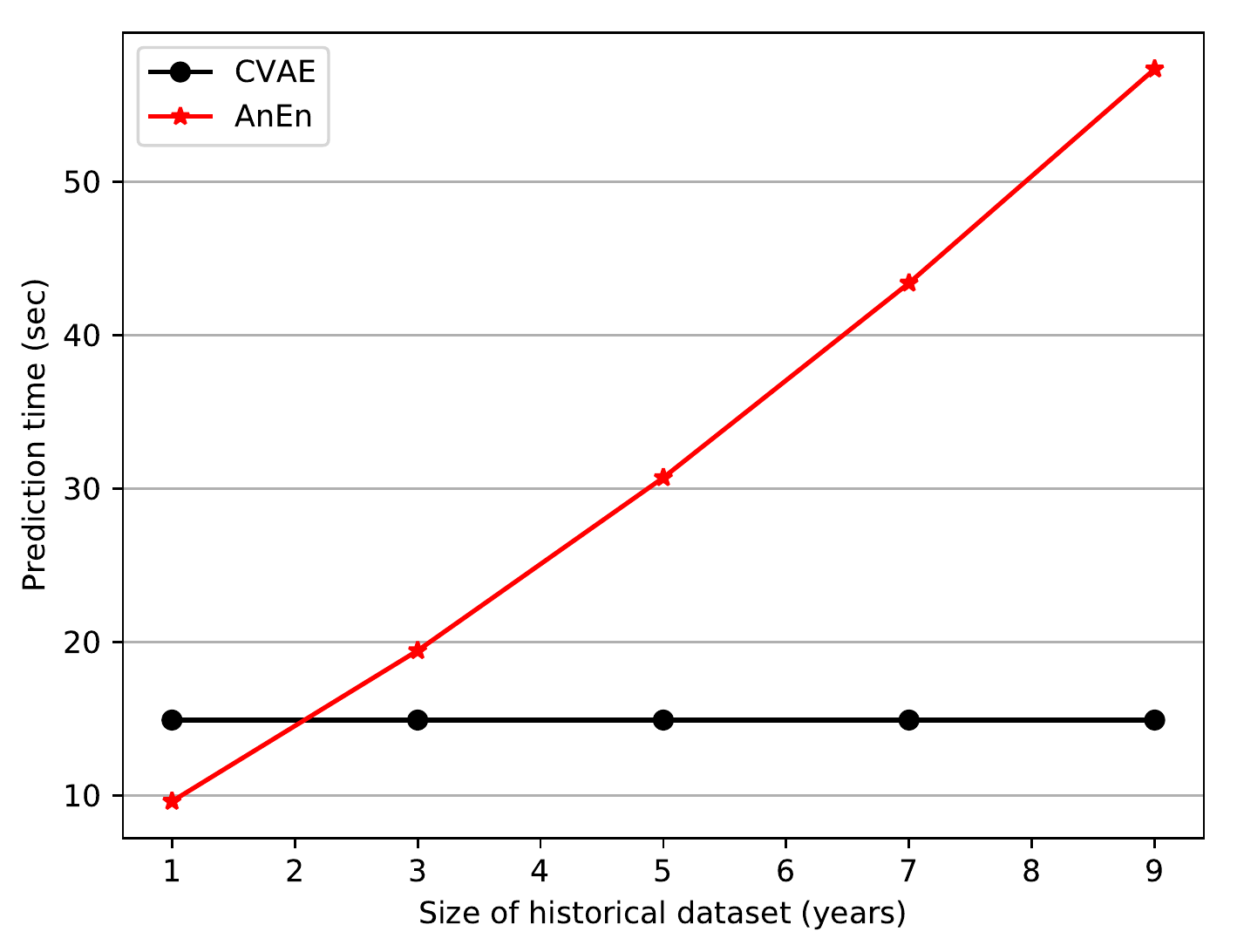}
\end{center}
   \caption{Runtime comparisons between CVAE and AnEn models}
\label{fig:timing}
\end{figure}

Using CVAE generally shows inferior results when compared to AnEn, but nevertheless within a small error margin.  Given the very large advantage of CVAE in terms of memory and computational complexity, the trade-off between these advantage and the larger error become advantageous for a large class of problems.

Figure \ref{fig:timing} shows the predicted runtime for CVAE and AnEn. The main advantage of CVAE becomes prominent with large datasets.  In fact, while AnEn needs to linearly scan the entire dataset every time a new set of ensemble needs to be generated, CVAE generates them nearly instantaneously once the model has been trained.  The training of the model is a computationally expensive operation, however it is performed only once. Therefore, for large datasets (in our experiments large corresponds to larger than roughly one year of data), or in cases when the analogs must be generated in real time, CVAE provides a faster and more efficient solution, with results that while are inferior of AnEn, are still very good and can be used in many problems to provide a measure of uncertainty.

\section{Conclusion}\label{sec:conclusion}

The use of specialized hardware able to perform very efficiently only specific tasks (e.g. deep learning)
seems to be the best way to face the challenges imposed by the exascale era.
Furthermore, the memory bottleneck on modern computing architectures is more than ever a limiting factor,
mostly because the rate of improvement in microprocessor speed exceed the rate of improvement in DRAM memory.
In order to take advantages of the new specialized hardware and limit the penalty imposed by memory speed, portions of current scientific codes should be rewritten and new algorithms should be developed.

In this work, we accomplished this by transforming a well established method for probabilistic forecasting, the Analog Ensemble method, from a traditional programming style to a deep learning approach.
The approach proposed in this work not only provides a more efficient way to perform probabilistic forecasting using the specialized hardware for deep learning, but also shows how general generative models can be used effectively to generate probabilistic forecasting. Both these contributions represent make this paper a unique and innovative work.

The results show that despite the better performance of the AnEn method, the Conditional Variational Autoencoder used in this work produces reliable probabilistic forecasts using a fraction of the time and memory needed by the AnEn.
Thanks to the low memory requirements and the advantage of using new specialized hardware, this new approach represents a valid option for edge-computing probabilistic forecasting. Such a new possibility provides new opportunities to a large amount of safety-critical applications in conditions where power is very limited and no internet connection is available (e.g. catastrophic scenarios).

In the future, we plan to explore other generative models (e.g. Deep Belief Networks, Generative Adversarial Networks) to perform better probabilistic forecasting.


\section{Acknowledgments}
We wish to thank Dr. S. Alessandrini for providing his code to generate the confidence intervals for the verification statistics. Furthermore, we thank Summer Internships in Parallel Computational Science (SIParCS) at National Center for Atmospheric Research (NCAR) supported by National Science Foundation (NSF) as B.R. was financially supported to work on this research. 

\section*{References}

\end{document}